\documentclass{article}
\usepackage{spconf,amsmath,graphicx}
\usepackage{color}
\usepackage{times}
\usepackage{epsfig}
\usepackage{graphicx}
\usepackage{amsmath}
\usepackage{amssymb}
\usepackage{amsthm}
\usepackage{mathrsfs}
\usepackage{amsfonts,amssymb}
\usepackage{booktabs}
\usepackage{threeparttable}
\usepackage{multicol}
\usepackage{multirow}
\usepackage{verbatim}
\usepackage{color}

\title{Bound Tightening Network for Robust Crowd Counting}

\name{Qiming Wu}
\address{Individual Contributor}

\begin{document}
\maketitle
\begin{abstract}
Crowd Counting is a fundamental topic, aiming to estimate the number of individuals in the crowded images or videos fed from surveillance cameras. Recent works focus on improving counting accuracy, while ignoring the certified robustness of counting models. In this paper, we propose a novel \textbf{Bound Tightening Network} (BTN) for Robust Crowd Counting. It consists of three parts: \textit{base model}, \textit{smooth regularization module} and \textit{certify bound module}. The core idea is to propagate the interval bound through the base model (certify bound module) and utilize the layer weights (smooth regularization module) to guide the network learning. Experiments on different benchmark datasets for counting demonstrate the effectiveness and efficiency of BTN.
\end{abstract}


\section{Introduction}
Crowd counting aims to estimate the total number of pedestrians in static images or dynamic videos. This task has drawn great attention because of its variety of applications in the real world. However, accurately counting people in the crowds is challenging due to diverse crowd distributions, severe occlusion and large-scale variations. With the rapid progress of deep neural networks (DNNs) \cite{ma2019bayesian,cheng2019learning,zhang2016single,amirgholipour2018ccnn,liu2018crowd}, the recent data-driven models have gained excellent counting performance. These methods can be roughly divided into three groups: directly applying object detector on the
input image \cite{leibe2005pedestrian, li2008estimating}, learning a mapping from patches to a number \cite{chan2008privacy, chan2009bayesian} and summing over the predicted density map \cite{wan2020modeling, zeng2017multi}. The mainstream focus in the counting area has been towards exploiting the advances in density map based methods due to the remarkable representation learning ability.

Despite the dramatic performance improvement in counting accuracy, few works \cite{wu2021towards, liu2021harnessing} have been devoted to the robustness of the models. Current estimation-based crowd counting
models are highly vulnerable to adversarial examples since small perturbation may lead to wrong predictions. Wu et al. \cite{wu2021towards} first propose
a systematic and practical method on the evaluation of the robustness in the counting area. They design adversarial patches to successfully attack several counting models in white-box and black-box forms. Further, Liu et al. \cite{liu2021harnessing} propose a 
Perceptual Adversarial Patch (PAP) framework, which promotes the
transferability of our adversarial patches by exploiting the model
scale and position perceptions. But they only focus on designing attack algorithms, leaving the defense pattern unresolved.

\begin{figure}
    \centering
    \includegraphics[width=0.46\textwidth]{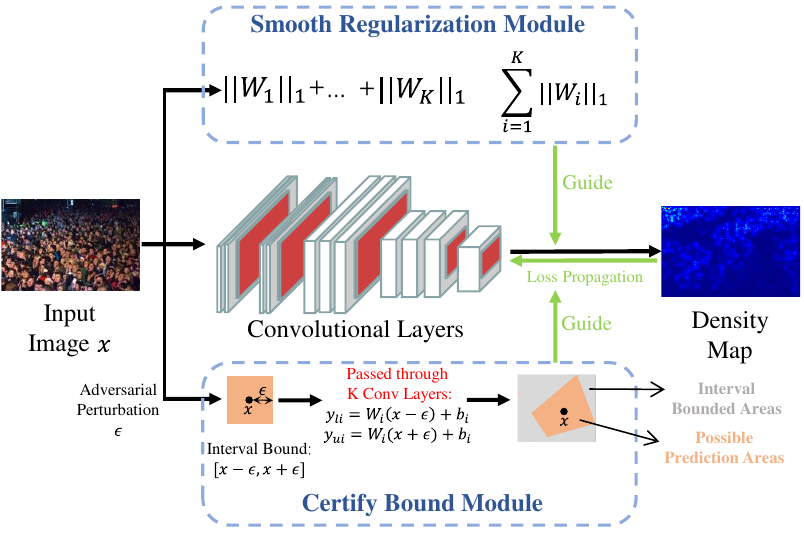}
    \caption{Overview of our proposed crowd counting model. It consists of three parts:\ \textit{Smooth Regularization Term}, \textit{Crowd Counting Base Model} and \textit{Certify Bound Module}.}
     \label{fig:illustration of reg-ibp}
     \vspace{-1em}
\end{figure}

In this paper, we propose a novel \textbf{Bound Tightening Network} (BTN) for Robust Crowd Counting. Specifically, the model is composed of three parts (as shown in Figure \ref{fig:illustration of reg-ibp}): \textit{Smooth Regularization Module}, \textit{Base Model} and \textit{Certify Bound Module}. In the smooth regularization module, BTN utilizes layer weights of the base model to introduce the regularization term in certificate model training. This module helps smoothen the training loop and benefits the final performances against adversarial examples. In the certify bound module, we manually introduce the adversarial perturbation $\epsilon$ and construct the initial interval bound $[x-\epsilon, x+\epsilon]$ based on the input image $x$. Then, the module propagates the interval bound through model layers and provides a possible prediction area to guide the later training loop. After iterations of certificate training, BTN finally becomes robust against adversarial perturbations and could provide a tight bound of the model prediction.

In summary, our contributions are as follows: we propose a novel network named BTN for robust crowd counting. It not only provides theoretical guarantees of the model prediction against adversarial examples, but also enhances the model robustness on different standard datasets in crowd counting.

\section{Methodology}

\textbf{Regression Based Crowd Analysis.} Given a set of $N$ labeled images $D=\{(x_i, l_i) \}_{i=1}^N$, where 
$x_i \in \mathbb{R}^{H_I \cdot W_I \cdot C_I}$ and $H_I$, $W_I$, and $C_I$ are the height, width, and channel number of the image, respectively.
$l_i$ is the ground truth density map of image $x_i$. 
Then, the crowd analysis task aims to learn a model $f_{\theta}$, parameterized by $\theta$, which can map from a crowd image to a corresponding density map by using these labeled images and solving the following optimization problem:
\begin{align}
   \min_{\theta} \frac{1}{2N} \sum^N_{i=1} \vert \vert f_\theta (x_i) - l_i \vert \vert^2.
\end{align}
Note that researchers recently have adopted more effective loss functions in crowd counting \cite{cheng2019learning,ma2019bayesian} and we consider the most commonly used $L_2$ loss function. Moreover, different crowd counting models will use different architectures. For instance, MCNN \cite{zhang2016single} uses Multi-column convolutional neural networks to predict the density map. The learned model $f_{\theta}$ can be used to predict the crowd count in a testing image $x$. Specifically, $f_{\theta}$ takes $x$ as an input and outputs the predicted density map $f_{\theta}(x_i)$. Then, the crowd count in $x$ is estimated by summing up all values of the density map.\par

\textbf{Threat Model.} In this paper, we focus on white-box adversarial attacks, which represent the most powerful adversary since it has access to the parameters and architecture of the target model. We now define the $L_p$ norm adversary formally as follows:

\textbf{($L_p$, $\epsilon$)-Adversary in Crowd Counting.} Given the dataset $D = \{(x_i, l_i) \in \mathcal{X} \times \mathcal{Y}|i=1, ..., N\}$, where $\mathcal{X}$, $\mathcal{Y}$ denotes the input space and the true label space, respectively. The Adversary will generate a well-crafted input $\tilde{x} \in B_{p,\epsilon}(x_i)$ such that $f_{\theta}(\tilde{x}) \neq l_i$ (with large distances).\par

\subsection{Certify Bound Module}
The certify bound module exams if the model output satisfies a given specification. On crowd counting models, we certify the pixels of the output density map, namely, every pixel value will be bounded within an interval:
\begin{align}
\begin{aligned}
    \forall x_0 & \in S(x_k, \epsilon) = \{x|||x-x_k||_{\infty} \le \epsilon\}, \\ & z_{Ki} \in [\underline{z}_{Ki}, \overline{z}_{Ki}],\label{equ:verification def}
\end{aligned}
\end{align}
where $z_{Ki}$ denotes the $i$-th pixel of the model output $z_{Ki}$ and $\underline{z}_{Ki}$, $\overline{z}_{Ki}$ represent the element-wise lower bound and upper bound for $z_{Ki}$. Now consider the Groundtruth map $GT$ for the $z_{Ki}$, $GT_i$ is the $i$-th pixel value, we have:
\begin{align}
    \mathop{min}_{\theta}
    \sum_i \mathop{max}(|GT_i - \underline{z}_{Ki}|,\ |\overline{z}_{Ki} - GT_i|),\label{equ:crowd counting verification}
\end{align}
where the inner $\sum_i \mathop{max}(|GT_i - \underline{z}_{Ki}|,\ |\overline{z}_{Ki} - GT_i|)$ denotes the verification goal, namely, finding the \textbf{worst-case} robustness bounds. Besides, the outer part $\mathop{min}_{\theta}$ means to turn the verification to the robustness optimization for a tighter bound by training the model $f_{\theta}$ through the verification equation.\par

\begin{table*}[htb]
\centering\setlength{\tabcolsep}{3.5mm}
\caption{Certified MAE and MSE results under $L_{\infty}$ circumstances on ShanghaiTech A and B datasets \cite{zhang2016single}. The numbers in the first line of the table represents: "1": clean trained models, "2": the best model of our certifiable training, "3": models trained without the "smooth regularization module", "4": models trained with the $L_2$ regularization term. Note that we train all the models for 400 epochs with the same strategy (e.g., same learning rate, batch size and so on.) for a fair comparison.}\label{tab: shanghai A L_inf}
\begin{tabular}{|ccc|cccc|cc|}
\hline
\multicolumn{3}{|c|}{\multirow{2}{*}{Dataset}}                                                                      & \multicolumn{4}{c|}{Shanghai A}                                                                                & \multicolumn{2}{c|}{Part B}                     \\ \cline{4-9} 
\multicolumn{3}{|c|}{}                                                                                              & \multicolumn{1}{c|}{1}        & \multicolumn{1}{c|}{2}               & \multicolumn{1}{c|}{3}        & 4       & \multicolumn{1}{c|}{1}        & 2               \\ \hline
\multicolumn{1}{|c|}{\multirow{2}{*}{$\epsilon=0$}}     & \multicolumn{1}{c|}{\multirow{2}{*}{clean test}}    & MAE & \multicolumn{1}{c|}{110.15}   & \multicolumn{1}{c|}{\textbf{120.61}} & \multicolumn{1}{c|}{126.72}   & 144.64  & \multicolumn{1}{c|}{25.21}    & \textbf{35.42}  \\ \cline{3-9} 
\multicolumn{1}{|c|}{}                                  & \multicolumn{1}{c|}{}                               & MSE & \multicolumn{1}{c|}{169.29}   & \multicolumn{1}{c|}{\textbf{183.74}} & \multicolumn{1}{c|}{197.18}   & 223.40  & \multicolumn{1}{c|}{43.68}    & \textbf{61.32}  \\ \hline
\multicolumn{1}{|c|}{\multirow{4}{*}{$\epsilon=1/255$}} & \multicolumn{1}{c|}{\multirow{2}{*}{certify-tight}} & MAE & \multicolumn{1}{c|}{2967.50}  & \multicolumn{1}{c|}{\textbf{157.24}} & \multicolumn{1}{c|}{1031.09}  & 772.39  & \multicolumn{1}{c|}{667.31}   & \textbf{39.25}  \\ \cline{3-9} 
\multicolumn{1}{|c|}{}                                  & \multicolumn{1}{c|}{}                               & MSE & \multicolumn{1}{c|}{3644.75}  & \multicolumn{1}{c|}{\textbf{219.37}} & \multicolumn{1}{c|}{1287.24}  & 937.56  & \multicolumn{1}{c|}{741.97}   & \textbf{64.40}  \\ \cline{2-9} 
\multicolumn{1}{|c|}{}                                  & \multicolumn{1}{c|}{\multirow{2}{*}{certify-pixel}} & MAE & \multicolumn{1}{c|}{3178.36}  & \multicolumn{1}{c|}{\textbf{467.97}} & \multicolumn{1}{c|}{1121.73}  & 1296.11 & \multicolumn{1}{c|}{766.46}   & \textbf{160.73} \\ \cline{3-9} 
\multicolumn{1}{|c|}{}                                  & \multicolumn{1}{c|}{}                               & MSE & \multicolumn{1}{c|}{3838.41}  & \multicolumn{1}{c|}{\textbf{553.43}} & \multicolumn{1}{c|}{1769.37}  & 2090.77 & \multicolumn{1}{c|}{845.64}   & \textbf{183.02} \\ \hline
\multicolumn{1}{|c|}{\multirow{4}{*}{$\epsilon=3/255$}} & \multicolumn{1}{c|}{\multirow{2}{*}{certify-tight}} & MAE & \multicolumn{1}{c|}{20883.33} & \multicolumn{1}{c|}{\textbf{235.54}} & \multicolumn{1}{c|}{5448.97}  & 3010.43 & \multicolumn{1}{c|}{5048.04}  & \textbf{46.91}  \\ \cline{3-9} 
\multicolumn{1}{|c|}{}                                  & \multicolumn{1}{c|}{}                               & MSE & \multicolumn{1}{c|}{24037.31} & \multicolumn{1}{c|}{\textbf{305.77}} & \multicolumn{1}{c|}{6433.75}  & 3498.44 & \multicolumn{1}{c|}{5300.03}  & \textbf{69.84}  \\ \cline{2-9} 
\multicolumn{1}{|c|}{}                                  & \multicolumn{1}{c|}{\multirow{2}{*}{certify-pixel}} & MAE & \multicolumn{1}{c|}{20897.64} & \multicolumn{1}{c|}{\textbf{561.93}} & \multicolumn{1}{c|}{5574.82}  & 3232.59 & \multicolumn{1}{c|}{5067.04}  & \textbf{168.89} \\ \cline{3-9} 
\multicolumn{1}{|c|}{}                                  & \multicolumn{1}{c|}{}                               & MSE & \multicolumn{1}{c|}{24049.13} & \multicolumn{1}{c|}{\textbf{660.71}} & \multicolumn{1}{c|}{6547.82}  & 3715.16 & \multicolumn{1}{c|}{5319.03}  & \textbf{191.39} \\ \hline
\multicolumn{1}{|c|}{\multirow{4}{*}{$\epsilon=5/255$}} & \multicolumn{1}{c|}{\multirow{2}{*}{certify-tight}} & MAE & \multicolumn{1}{c|}{49628.48} & \multicolumn{1}{c|}{\textbf{326.56}} & \multicolumn{1}{c|}{12842.08} & 6054.21 & \multicolumn{1}{c|}{11537.62} & \textbf{54.61}  \\ \cline{3-9} 
\multicolumn{1}{|c|}{}                                  & \multicolumn{1}{c|}{}                               & MSE & \multicolumn{1}{c|}{55631.89} & \multicolumn{1}{c|}{\textbf{417.21}} & \multicolumn{1}{c|}{14651.14} & 6900.38 & \multicolumn{1}{c|}{11877.07} & \textbf{75.92}  \\ \cline{2-9} 
\multicolumn{1}{|c|}{}                                  & \multicolumn{1}{c|}{\multirow{2}{*}{certify-pixel}} & MAE & \multicolumn{1}{c|}{49629.88} & \multicolumn{1}{c|}{\textbf{673.04}} & \multicolumn{1}{c|}{12874.49} & 6159.22 & \multicolumn{1}{c|}{11542.33} & \textbf{177.57} \\ \cline{3-9} 
\multicolumn{1}{|c|}{}                                  & \multicolumn{1}{c|}{}                               & MSE & \multicolumn{1}{c|}{55633.08} & \multicolumn{1}{c|}{\textbf{790.27}} & \multicolumn{1}{c|}{14679.88} & 7001.09 & \multicolumn{1}{c|}{11881.82} & \textbf{200.28} \\ \hline
\end{tabular}
\end{table*}

\subsection{Smooth Regularization Module}
To begin with, we first propose the \emph{Norm Duality} theorem based on the \emph{H$\ddot{o}$lder Inequality} to theoretically guarantee the legitimacy of using L1 and L2 regularization in the robustness training process.

\newtheorem{thm}{Theorem }
\begin{thm}\label{theorem:norm duality}\textbf{(Norm Duality)}
Consider the activation function $ReLU$, the model $f$ have $K$ affine layers, $z_0 = x$ and $W_i$ represents the $i$-th affine layer. Given the adversarial example $\{\tilde{x}: ||\tilde{x} - x||_{\infty} \le \epsilon_1\}$ as the input example, we have:
\begin{align}
    ||f(\tilde{x})-f(x)||_{\infty} \le \epsilon_1\prod_{m=1}^{K}\mathop{max}_{j}||W_{mj}||_1.
\end{align}
\end{thm}

Based on the Theorem \ref{theorem:norm duality}, we now optimize the right part as to verifiably train the model. For total $K$ affine layers, we modify the optimization part as $\mathop{min}_{\theta} \sum_{i=1}^K \vert\vert W_i \vert\vert_1$, which is the $L1$ regularization form. Further more, we derive the following lemma:

\newtheorem{lemma}{Lemma}
\begin{lemma}\label{lemma: lemma 1.}($L_2$ Norm Cases). Consider the $L_2$ norm bounded adversarial example $\{\tilde{x}: ||\tilde{x}-x||_2 \le \epsilon_2 \}$ and the notations in Theorem.\ref{theorem:norm duality}, for any neuron $j$ of $z_1$, we have:
\begin{align}
    |\tilde{z}_{1j}-z_{1j}|  \le \epsilon_2 ||W_{1j}||_2 .
\end{align}
\end{lemma}
\begin{proof}
The detailed proof of the theorem and the lemma will be shown in the code repositories.
\end{proof}

Considering Theorem \ref{theorem:norm duality} and Lemma \ref{lemma: lemma 1.}, our loss is composed of three parts: \emph{Normal training loss}, \emph{Certify error loss} and \emph{Regularization loss}. The general form of the practical loss is:
\begin{align}
    \mathcal{L}_{total} = \kappa L_{normal} + (1 - \kappa) L_{certify} + \lambda L_{reg}.
\end{align}

For regression-based crowd counting models, we use the \emph{Mean Squared Error} (MSE) loss as it is the most widely used loss form in the field of crowd counting. Then we rewrite the total loss function $\mathcal{L}_{total}$:
\begin{align}
\begin{aligned}
     & \mathcal{L}_{total} = \frac{1}{2N} \sum_{j=1}^N \{ \underbrace{\kappa ||f_{\theta}(x_j) - l_j||^2_{2}}_{\text{natural \ loss}} \\ & + \underbrace{(1-\kappa) \sum_i [\mathop{max}(|GT_i - \underline{z}_{Ki}|, |\overline{z}_{Ki} - GT_i|)]^2}_{\text{certify\ error \ loss}} \\ & + \underbrace{\lambda \sum_{m=1}^K ||W_m||_1}_{\text{regularization\ loss}}\},\label{loss:MCNN train}
    \end{aligned}
\end{align}
where the certify error loss part aims to find the worst case robustness bound for the $i$-th pixel of the $j$-th image in the dataset. As we have discussed in Theorem.\ref{theorem:norm duality}, we replace the $\sum_{i=1}^K\mathop{max}_j||W_{ij}||_1$ with the $\sum_m^K ||W_m||_1$, which means to sum up the weights of the entire model layers to calculate the $L_1$ regularization loss.\par

\subsection{Robustness Optimization.}
We utilize above two modules to guide the robustness optimization (i.e., BTN certificate training). Given the base model $f$ initialized with $\theta_0$, we adopt the multi-stage training schedule: 50 epochs to warm up ($\kappa=1$ in Eq.\ref{loss:MCNN train}), 150 epochs for the slow decrease of $\kappa$ (from 1 to 0.5) and 200 epochs for the certificate training ($\kappa = 0.5$). Besides, the weight decay parameter $\lambda=1\times10^{-3}$ and $\beta = 10$ ($L_2$ cases).

\section{Experiments}
\subsection{Experiment Setup}
\textbf{Dataset.} For crowd counting, we verifiably train and evaluate the robustness bounds on the ShanghaiTech A and B dataset \cite{zhang2016single}, which is the most representative dataset in the field contains 1198 images with over 330,000 people annotated. For ablation studies on image classifiers, we select the most commonly used CIFAR10 and Tiny-ImageNet \cite{Acemap.176013195}.

\textbf{Model Structure.} In this paper, we utilize the MCNN \cite{zhang2016single} as the base model, which has three branches composed of convolutional and max-pooling layers. MCNN is popular for its great transferability to various tasks. For ablation studies on image classifiers, we use the same Medium and Large CNN models in IBP \cite{gowal2018effectiveness} for a fair comparison. And we use the same model as in "BCP" method \cite{lee2020lipschitz} to train on the Tiny-ImageNet \cite{Acemap.176013195}.\par

\textbf{Evaluation Metrics.} For crowd counting, we propose two metrics for the evaluation of robustness: \emph{certify-tight MAE \& MSE} and \emph{certify-pixel MAE \& MSE}:
\begin{align}
    \begin{aligned}
    & \text{Certify-tight\ MAE} = \frac{1}{N} \cdot \sum_{i=1}^{N} \vert C^{GT}_i-\tilde{C}_i \vert, \\ & \text{Certify-tight\ MSE} = \sqrt{\frac{1}{N} \cdot \sum_{i=1}^{N}(C^{GT}_i-\tilde{C}_i)^{2}},\label{equ:certify tight}
    \end{aligned}
\end{align}
where $N$ is the number of images, $C^{GT}_i$ denotes the $i$-th ground-truth counting and $\tilde{C}_i$ is the counting of the robustness bound (upper or lower bound). In fact, we select the maximum distance between $GT$ and the robustness bound, namely, $C^{GT}_i-\tilde{C}_i = max(|C^{GT}_i-C(\underline{z}_{Ki})|,\ |C(\overline{z}_{Ki}) - C^{GT}_i|)$. To better evaluate the worst case of every pixel on the output density map (measure the model performances on the pixel level), we introduce the \emph{Certify-pixel MAE \& MSE}:
\begin{align}
    \begin{aligned}
    & \text{\textbf{Certify-pixel MAE:}}\\ & \frac{1}{N}\sum_i^{N}\sum_j \mathop{max}(|\overline{z}_{Kij} - GT_{ij}|, |GT_{ij} - \underline{z}_{Kij}|), \\ & \text{\textbf{Certify-pixel MSE:}}\\ & \sqrt{\frac{1}{N}\sum_{i}^{N}\sum_j [\mathop{max}(|\overline{z}_{Kij} - GT_{ij}|, |GT_{ij} - \underline{z}_{Kij}|)]^2}.\label{equ:certify pixel}
    \end{aligned}
\end{align}

\begin{table*}[htb]
\centering
\setlength{\tabcolsep}{2.4mm}{
\caption{The comparison results of $L_2$ robustness bound by BTN models. Note that model "1" is clean-trained model and model "2" is the certifiably trained one. Also, the experiments is done on the ShanghaiTech A dataset \cite{zhang2016single} and we adopt the "certify-tight MAE $\&$ MSE" as the evaluation metric. The perturbation $\epsilon=0, 0.5, 1.0, 2.0$.}\label{tab:L2 shanghai A}
\begin{tabular}{|c|cc|cc|cc|cc|}
\hline
\multirow{2}{*}{Model} & \multicolumn{2}{c|}{Epsilon=0}       & \multicolumn{2}{c|}{Epsilon=0.5}                       & \multicolumn{2}{c|}{Epsilon=1.0}                       & \multicolumn{2}{c|}{Epsilon=2.0}                       \\ \cline{2-9} 
                       & \multicolumn{1}{c|}{MAE}    & MSE    & \multicolumn{1}{c|}{MAE}             & MSE             & \multicolumn{1}{c|}{MAE}             & MSE             & \multicolumn{1}{c|}{MAE}             & MSE             \\ \hline
1                      & \multicolumn{1}{c|}{110.15} & 169.29 & \multicolumn{1}{c|}{1572372.40}      & 1759524.73      & \multicolumn{1}{c|}{3528835.11}      & 3931424.35      & \multicolumn{1}{c|}{7476160.77}      & 8266575.34      \\ \hline
2                      & \multicolumn{1}{c|}{150.23} & 223.24 & \multicolumn{1}{c|}{\textbf{355.75}} & \textbf{443.54} & \multicolumn{1}{c|}{\textbf{550.83}} & \textbf{643.72} & \multicolumn{1}{c|}{\textbf{697.94}} & \textbf{812.66} \\ \hline
\end{tabular}}
\end{table*}

Our results of BTN on ShanghaiTech part A $\&$ B datasets \cite{zhang2016single} under $L_{\infty}$ circumstances are summarized in Table \ref{tab: shanghai A L_inf}. From the row of $\epsilon=0$ we can test the clean error of verifiably trained BTN. Our verifiably trained networks only increase 10.46 on MAE on ShanghaiTech A dataset and 10.21 on B dataset \cite{zhang2016single}. One of the disadvantages of certificate models is the high error on clean examples, which leads to the limited applications of previous works. However, the clean error of our proposed BTN is acceptable. Apart from that, by comparing the certified results between clean trained and verifiably trained models (column "1" and "2" of "ShanghaiTech A" and "ShanghaiTech B" in Table \ref{tab: shanghai A L_inf}), it is obvious that our verifiably trained models achieve much tighter robustness bound.\par

\begin{table}
\centering\setlength{\tabcolsep}{1.5mm}
\caption{Ablation study on image classifications. We use standard and verification Error to evaluate on CIFAR10 and Tiny-ImageNet dataset. "Std" and "Ver" denote the \emph{standard error} and \emph{verified error}. Note that the experimental data of IBP and CROWN-IBP are reported in Table 2 of \cite{zhang2019towards}, which has been widely recognized to be reproducible.}\label{tab:CIFAR and tiny imagenet results}
\begin{tabular}{|c|cccc|cc|}
\hline
\multirow{3}{*}{Method}                             & \multicolumn{4}{c|}{CIFAR-10}                                                                                                    & \multicolumn{2}{c|}{Tiny-ImageNet}                   \\ \cline{2-7} 
                                                    & \multicolumn{2}{c|}{$\epsilon=8/255$}                                     & \multicolumn{2}{c|}{$\epsilon=16/255$}               & \multicolumn{2}{c|}{$\epsilon=1/255$}                \\ \cline{2-7} 
                                                    & \multicolumn{1}{c|}{Std}            & \multicolumn{1}{c|}{Ver}            & \multicolumn{1}{c|}{Std}            & Ver            & \multicolumn{1}{c|}{Std}            & Ver            \\ \hline
IBP                                                 & \multicolumn{1}{c|}{58.43}          & \multicolumn{1}{c|}{70.81}          & \multicolumn{1}{c|}{68.97}          & 78.12          & \multicolumn{1}{c|}{-}              & -              \\ \hline
\begin{tabular}[c]{@{}c@{}}CROWN\\ IBP\end{tabular} & \multicolumn{1}{c|}{54.02}          & \multicolumn{1}{c|}{66.94}          & \multicolumn{1}{c|}{66.06}          & 76.80          & \multicolumn{1}{c|}{-}              & -              \\ \hline
Ours                                                & \multicolumn{1}{c|}{\textbf{52.93}} & \multicolumn{1}{c|}{\textbf{62.83}} & \multicolumn{1}{c|}{\textbf{58.89}} & \textbf{69.19} & \multicolumn{1}{c|}{\textbf{83.52}} & \textbf{88.78} \\ \hline
\end{tabular}
\end{table}

\textbf{Ablation Study of BTN on Crowd Counting Models.} To better clarify the effectiveness of the proposed BTN module, we also study the cases of certificate training without $L_1$ regularization loss and training with $L_2$ regularization term (column "3" and "4" of "ShanghaiTech A" in Table \ref{tab: shanghai A L_inf}). We discover that neither the robustness bound of column "3" nor that of column "4" is tighter than column "2" (our proposed BTN). This phenomenon demonstrates that in the smooth regularization module the regularization term is theoretically guaranteed, other regularization terms will decrease the effectiveness of BTN.

\textbf{BTN under $L_2$ Circumstances.} Besides the $L_{\infty}$ norm cases, we also study the verifiable robustness of BTN under $L_2$ circumstances. The results are summarized in Table \ref{tab:L2 shanghai A}. We compare the tightness of robustness between the clean-trained and verifiably trained networks. Although the clean error is slightly higher than that of $L_{\infty}$ cases, it is still acceptable (namely, clean MAE is increased by 40.08 and clean MSE 53.95). For the most common used $L_2$ norm perturbation $\epsilon=0.5,\ 1.0,\ 2.0$, BTN performs quite well. As can be seen in Table \ref{tab:L2 shanghai A}, the \emph{certify-tight} MAE and MSE values of the clean-trained model is much larger than those of our method trained models (e.g., when $\epsilon=0.5$, the clean MAE value is 4419.88 times larger than the verifiable MAE value). Moreover, we discover that the gap of the tightness of $L_2$ robustness bound is increasingly enlarged with the $\epsilon$ increasing. This phenomenon highlights the need of certified defense of regression models against $L_2$ attacks.\par

\textbf{Ablation Study on Classifications.} To better clarify our work, we implement BTN on the popular image classification dataset: CIFAR10 and Tiny-ImageNet \cite{Acemap.176013195}. Particularly, the implementation time of BTN is comparable with that of IBP. The results are reported in Table \ref{tab:CIFAR and tiny imagenet results}. We train the "large CNN" model used in IBP \cite{gowal2018effectiveness} on $\epsilon=8/255,\ 16/255$ and test the best models respectively. In Table \ref{tab:CIFAR and tiny imagenet results}, we achieve comparable clean and verification error compared with CROWN-IBP \cite{zhang2019towards} and the verification error is lower than IBP results for 2.74\%. For $\epsilon = 8/255,\ 16/255$, our method defeats IBP and CROWN-IBP on both clean and verification errors. We contribute the success of BTN to the smoothness of the initial training stage. Moreover, we also do the $L_2$ norm case study. We experiment BTN on CIFAR10 dataset in $L_2$ norm ($\epsilon=0.14$). We achieve the verification error as 85.08\%.

\section{Conclusion and future work}
We creatively propose the efficient and scalable regression-based neural network certification method named BTN, which adopts the "smooth regularization module" and the "certify bound module" in the certificate training loops. Moreover, we theoretically demonstrate the feasibility and the tight bound of BTN, and experiment with BTN on the popular crowd counting dataset along with some classification datasets. We find that BTN performs effectively and efficiently on these datasets and even better than the state-of-the-art method like CROWN-IBP. In addition, BTN has strong practicability and can be compatible with the mainstream crowd counting networks as basic models. Our work highlights the need for future works on the verification of neural networks.\par

\vfill\pagebreak
\bibliographystyle{IEEEbib}
\bibliography{ref}
\end{document}